\documentclass[runningheads]{llncs}

\usepackage[mobile]{eccv}

\usepackage{eccvabbrv}

\usepackage{graphicx}
\usepackage{booktabs}

\usepackage[accsupp]{axessibility}  

\usepackage[pagebackref,breaklinks,colorlinks,citecolor=eccvblue,linkcolor=eccvblue]{hyperref}

\usepackage{orcidlink}

\usepackage{multirow}
\usepackage{array}
\usepackage{diagbox}
\usepackage{amsmath}
\usepackage{multibib}

\newcommand{\rb}[1]{\textbf{{\color{red}#1}}}
\newcommand{\bu}[1]{\underline{{\color{blue}#1}}}
\newcommand{\modelname}{\textit{AdaIR}}

\newcolumntype{P}[1]{>{\centering\arraybackslash}p{#1}}

\begin{document}

\title{AdaIR: Exploiting Underlying Similarities of\\Image Restoration Tasks with Adapters} 

\titlerunning{AdaIR}
\authorrunning{H.-W. Chen, Y.-S. Xu, K.-C.K. Chan \etal}

\author{
Hao-Wei Chen\textsuperscript{1} \ Yu-Syuan Xu\textsuperscript{2} \ Kelvin C.K. Chan\textsuperscript{3} \ Hsien-Kai Kuo\textsuperscript{2} \\ Chun-Yi Lee\textsuperscript{1} \ Ming-Hsuan Yang\textsuperscript{4}
\smallskip
\\
\textsuperscript{1}National Tsing Hua University~~~ \textsuperscript{2}MediaTek Inc.\\
\textsuperscript{3}Google~~~ \textsuperscript{4} University of California, Merced \\
}
\institute{}



\maketitle

\begin{abstract}
Existing image restoration approaches typically employ extensive networks specifically trained for designated degradations.
Despite being effective, such methods inevitably entail considerable storage costs and computational overheads due to the reliance on task-specific networks. 
In this work, we go beyond this well-established framework and exploit the inherent commonalities among image restoration tasks. 
The primary objective is to identify components that are shareable across restoration tasks and augment the shared components with modules specifically trained for individual tasks. 
Towards this goal, we propose \modelname, a novel framework that enables low storage cost and efficient training without sacrificing performance. Specifically, a generic restoration network is first constructed through self-supervised pre-training using synthetic degradations. 
Subsequent to the pre-training phase, adapters are trained to adapt the pre-trained network to specific degradations. \modelname~requires solely the training of lightweight, task-specific modules, ensuring a more efficient storage and training regimen. 
We have conducted extensive experiments to validate the effectiveness of \modelname~and analyze the influence of the pre-training strategy on discovering shareable components.
Extensive experimental results show that \modelname~achieve outstanding results on multi-task restoration while utilizing significantly fewer parameters (1.9 MB) and less training time (7 hours) for each restoration task.
The source codes and trained models will be released. 
\end{abstract}    

\section{Introduction}
\label{sec::introduction}

\begin{figure*}[t]
    \centering
    \includegraphics[width=1\textwidth]{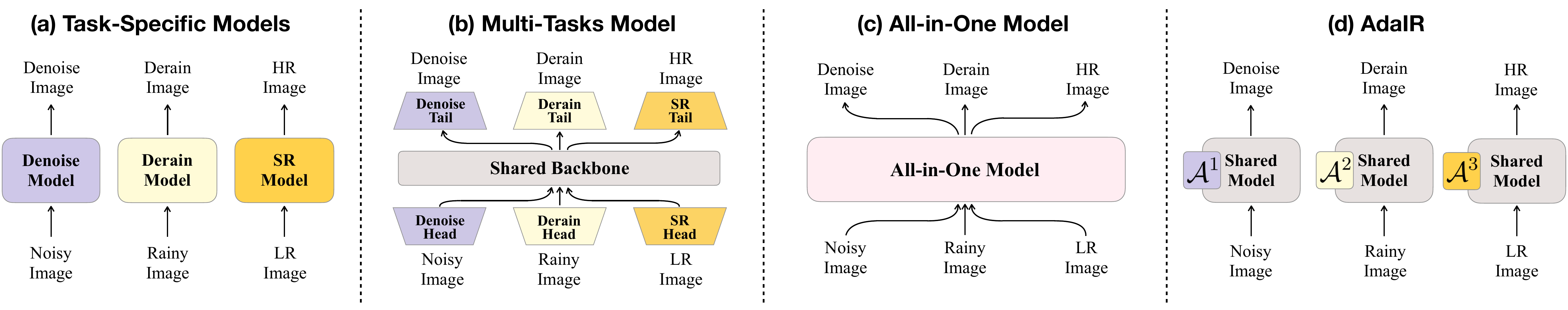}
    \caption{\textbf{Comparison of different approaches.} An illustration of different strategies to deal with multiple image restoration tasks. (a) Training separate models for each specific task, \eg, denoising, deraining, and supre-resolution. (b) Appending multiple heads and tails, which are respectively tailored to different tasks, on a shared backbone model. (c) Designing special blocks within the all-in-one model to encode and utilize degradation information without specifying the task explicitly. (d) Our proposed \modelname~exploiting task-specific adapters to address different restoration tasks.}
    \label{fig::teaser}
\end{figure*}

Image restoration is a long-standing problem in low-level vision, with the objective of reconstructing high-quality (HQ) images from degraded low-quality (LQ) counterparts. This field has witnessed substantial progress with the emergence of deep learning approaches. Current methods achieve considerable success by training models tailored for a specific degradation, as presented in Fig.~\ref{fig::teaser}(a). Although the single-task approaches have achieved significant success in a wide range of tasks, they are well-known to be confined to the degradations present during the training phase, resulting in limited generalizability. 

While image restoration covers a spectrum of degradation types, the primary objective remains to eliminate artifacts caused by degradations and recover a high-quality image. By understanding learned latent representations of different degradations, it is discovered that there are intersecting latent representations across different tasks~\cite{idr, airnet}. Therefore, a flexible and cost-efficient approach to image restoration can be realized through a common foundation complemented by compact, task-specific modules. Straightforward strategies for multi-task image restoration involve directly  training a shared backbone Fig.~\ref{fig::teaser}(b) or a all-in-one model Fig.~\ref{fig::teaser}(c) on multiple degradations. These methods enhance the versatility of the trained network and reduces the storage costs of multiple single-task models. However, they still exhibit limited generalizability to degradations beyond those included in the training set, ultimately constraining the scalability of existing methods.


In this work, we explore and exploit the shareable components among different restoration tasks with parameter-efficient tuning. This approach divides the training process into pre-training and fine-tuning phases. The pre-training phase aims to uncover shareable components, while the fine-tuning phase facilitates easy adaptation to different tasks. Although parameter-efficient tuning has been extensively studied in natural language processing (NLP)~\cite{brown2020, prefix_tuning, prompt_tuning, residual_adapter, adapter, adapterscale, adapterfusion, adaptershare, mam_adapter, compacter, lora, bitfit} and high-level vision tasks~\cite{zhou2022, maple, vpt, adaptformer, fact, petl_vit, glora}, its application in the low-level vision domain remains under explored. 
The benefits of parameter-efficient tuning are twofold. Firstly, the relationship between various restoration tasks is difficult to discern when training a single multi-task model from scratch. The two-phase transfer learning mechanism allows for learning shareable components during the pre-training phase. This allows us to analyze pre-training schemes instead of training restoration models from scratch, further investigating generalizability. Secondly, efficient fine-tuning enables the lightweight task-specific extensions of the model to address unseen degradations not covered in the pre-training phase, thereby reducing memory and computational time. 
Combining the above insights and the merits of parameter-efficient tuning, we aim to advance research this approach for multi-task image restoration. To achieve this, we design a novel adapter module as our parameter-efficient tuning method.

We introduce \modelname, a framework that leverages adapters for efficient adaptation to previously unseen degradations through pre-training and fine-tuning phases. Our framework initiates a self-supervised pre-training phase that employs synthetic degradations. To integrate new tasks, lightweight adapters are inserted into the foundation model. During the fine-tuning phase, the pre-trained model remains unchanged, and only the adapters undergo training. Our approach enables the utilization of a shared foundation model across various tasks, resulting in reduced training time compared to training from scratch. Additionally, we only need to store adapters for different tasks rather than multiple models, thereby reducing storage costs. 

%

As our objective is to investigate the how to explore shareable components across diverse restoration tasks, our focus is on the effect of various pre-training schemes employed during the pre-training phase on the performance of downstream tasks in the fine-tuning phase. We perform multiple experiments with various pre-training schemes tailored to the downstream tasks. The extensive studies on pre-training schemes not only assist us in training a  generic model but also provide valuable insights for future advancements. 

Moreover, we provide comparisons with existing multi-task restoration methods, which reveals that \modelname~could achieve comparable performance against existing works with lightweight parameters and corresponding training strategy, thereby confirming the efficacy of our approach. 

The main contributions of our work are:
\begin{itemize}
    \item We propose \modelname, a framework that integrates adapter modules into a shareable foundation model for efficient adaptation to novel restoration tasks.
    \item Our detailed analysis studies the influence of the pre-training strategy and adapter modules on the performance, offering guidance for future research.
    \item Extensive experiments demonstrate that \modelname~achieves favorable performance with efficiency on various restoration tasks.
\end{itemize}

\section{Related Work}
\label{sec::related_work}
\subsubsection{Single-Task Restoration.} Existing image restoration methods mainly focus on the setting of a single task to recover high-quality images from low-quality images that endured specific degradation. Numerous restoration tasks benefit from the emergence of a deep learning-based model and achieve huge improvement. These tasks involve super-resolution (SR)~\cite{srcnn, vdsr, edsr, srresnet, esrgan, rcan, rdn, san, esrt, hat}, denoising~\cite{cbm3d, dncnn, ircnn, ffdnet, brdnet}, deraining~\cite{derainnet, didmdn, rescan, sirr, umrl, prene, mspfn, lpnet}, dehazing~\cite{dehazenet, mscnn, aodnet, epdn, griddehazenet, fdgan, ffanet, msbdn, kddn, aecrnet}, and deblurring~\cite{deepdeblur, deblurgan, srn, deblurganv2, gao2019, dmphn, mtrnn, dbgan, suin2020, msdinet, rsblur}.

\subsubsection{Multi-Task Restoration.} Recently, image restoration has developed toward extending the restoration model to multiple-task restoration. Some approaches introduce unified model architecture to address various types of degradation, such as HINet~\cite{hinet}, MPRNet~\cite{mprnet}, SwinIR~\cite{swinir}, Uformer~\cite{uformer}, and Restormer~\cite{restormer}. 
Despite significant success, these methods are still limited to applying a single model to multiple degradations. Each restoration model is typically trained for each specific degeneration with the same architecture. However, the requirement of training and storing several copies of the restoration model incurs extra memory and time consumption. 
Alternatively, several approaches~\cite {li2020, ipt, dl, transweather, airnet, vrdir, prores, promptir} propose to remove different degradations with a single model. 
IPT~\cite{ipt} introduces several head and tail modules into a shared backbone as demonstrated in Fig.~\ref{fig::teaser}(b), where each pair of head and tail is responsible for one task. AirNet~\cite{airnet} projects degenerated input images into latent space and utilizes contrastive learning to separate different degradations. Then, these features guide the restoration toward the specific tasks. PromptIR~\cite{promptir} proposes a prompt block to encode degradation-specific knowledge to lead the restoration model to multiple tasks. 
Despite these advancements, numerous challenges remain as these methods require retraining when encountering new types of image degradation. 
To tackle this issue, our approach involves pre-training a compact model designed to adapt to any image degradation via parameter-efficient tuning.

\subsubsection{Parameter-Efficient Tuning.} In the Natural Language Process (NLP) domain, transferring a large-scale pre-trained model, such as the Transformer-based~\cite{transformer} model, to multiple downstream tasks is now prevailing. The most intuitive manner is to fine-tune all the parameters in the pre-trained model to the specific task. However, this kind of approach inevitably entails the demands for additional memory and training costs. 
Thus, methods that efficiently adapt the pre-trained model to various tasks are proposed to mitigate the problem. These efficient tuning methods could be roughly categorized into three groups: (1) Prompt-Tuning~\cite{brown2020, prefix_tuning, prompt_tuning}, (2) Adapter~\cite{residual_adapter, adapter, adapterscale, adapterfusion, adaptershare, mam_adapter, compacter}, and (3) Low-Rank Adaptation (LoRA)~\cite{lora}. Furthermore, some works attempt to apply parameter-efficient tuning techniques on the vision tasks~\cite{zhou2022, maple, vpt, adaptformer, fact, petl_vit, glora}. 
In this work, we develop a novel adapter optimized for image restoration to deal with multi-task restoration effectively.

\section{Methodology}
\label{sec::methodology}
This section provides an overview of our proposed \modelname~framework, followed by the details of adapter modules within the framework. We then further discuss our training scheme, including pre-training and fine-tuning phases. 

\begin{figure*}[t]
    \centering
    \includegraphics[width=1\textwidth]{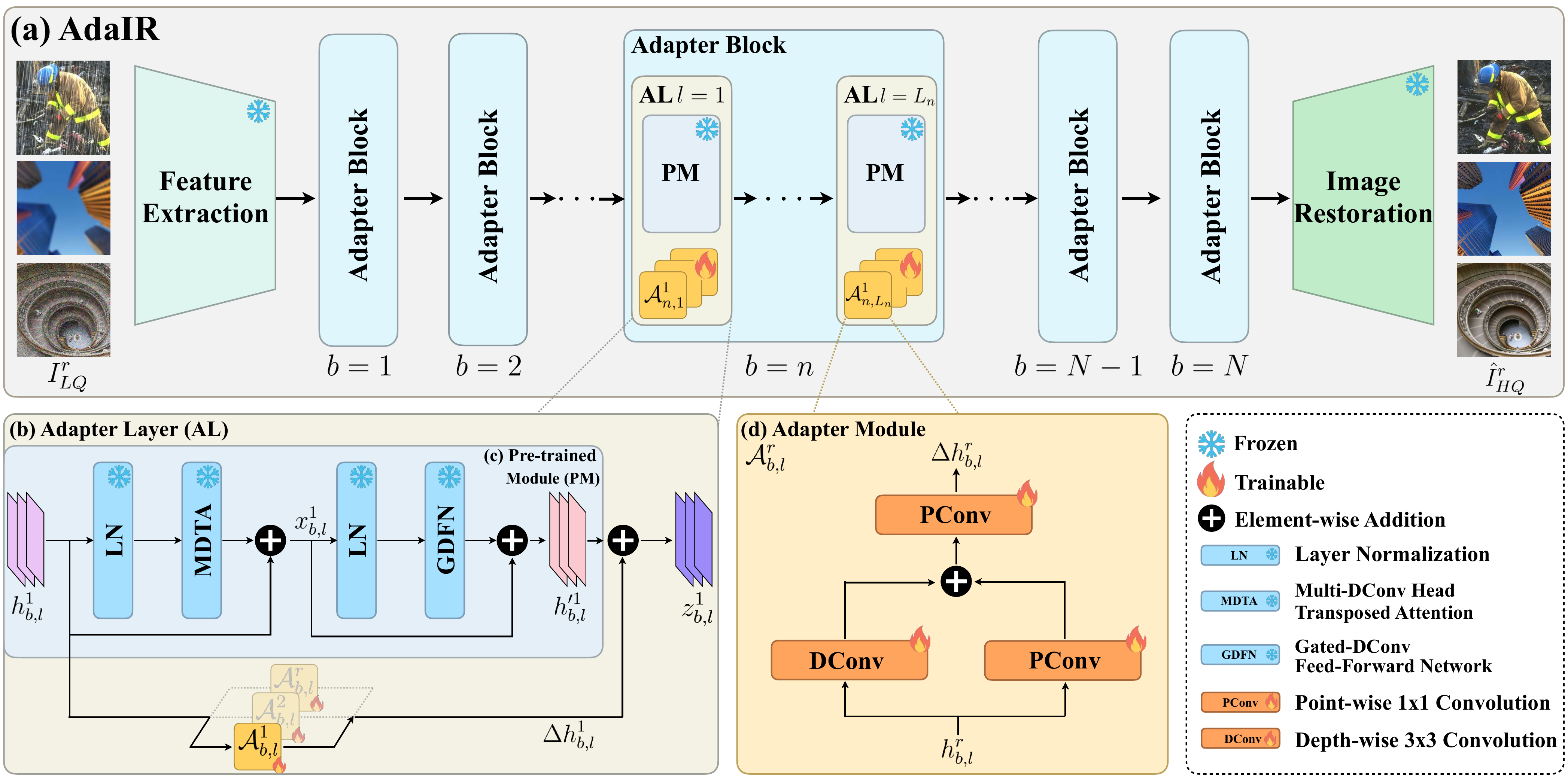}
    \caption{\textbf{Overview.} The proposed \modelname~framework. Our foundation model comprises feature extraction, pre-trained, and image restoration modules. The adapter modules interact with pre-trained modules to form adapter layers. When fine-tuning, the parameters of the foundation model are frozen; only the parameters in the adapter module are tunable.}
    \label{fig::overview}
\end{figure*}

\subsection{AdaIR}
\label{subsec::adair}
\modelname~is a framework designed to integrate adapter modules $\mathcal{A}^{R}$ into a foundation model, thereby enabling the generalization of the foundation model to multiple specific image restoration tasks. Fig.~\ref{fig::overview} presents an overview of the proposed \modelname~framework, which focuses on recovering high-quality (HQ) images $I^{r}_{HQ} \in \mathbb{R}^{H \times W \times 3}$ from the low-quality images (LQ) $I^{r}_{LQ} \in \mathbb{R}^{H \times W \times 3}$, where $r \in \{1,2,3, \ldots, R\}$ denotes the index of the restoration task and $R$ represents the total number of restoration tasks, with $H$ and $W$ referring to the height and width of the images, respectively. \modelname~tunes lightweight adapter modules $\mathcal{A}^{r}$ to learn task-specific knowledge for restoring $I^{r}_{HQ}$ by eliminating the artifacts present in the $I^{r}_{LQ}$. While keeping most of the foundation model's parameters frozen, we can efficiently address several restoration tasks without needing multiple separate restoration models.

\subsection{Instantiations}
\label{subsec::instantiations}
\subsubsection{Backbone.} In \modelname, the foundation model is designed to be flexible, which allows for the substitution of different underlying architectures. We have selected the architecture proposed in Restormer~\cite{restormer} as our foundation model for this study. Restormer is a unified architecture suitable for various image restoration tasks, which effectively demonstrate the capabilities of our proposed framework. It features a multi-level hierarchical encoder-decoder structure. The input image is initially projected into a latent embedding space through a feature extraction module, such as a 3x3 convolutional layer. Each level of the hierarchical architecture comprises several transformer blocks that serve as pre-trained modules. The Restormer progressively reduces the spatial resolution in the encoder. Conversely, the decoder incrementally upscales the spatial resolution until it matches the input image's. Ultimately, a convolutional layer processes the refined features to generate the residual image within the image restoration module. This residual image is then added to the input to produce the final restored image.

\subsubsection{Adapter Module.} To facilitate the learning of task-specific knowledge while leveraging a shareable foundation model, adapter modules are introduced, which play a crucial role in generalizing the foundation model to multiple restoration tasks. Our approach incorporates adapters into a sequence of layers as~\cite{adapter, mam_adapter, adaptformer}. We design an adapter architecture tailored for image restoration, as illustrated in Fig.~\ref{fig::overview}(d). Considering the characteristics of image restoration tasks in the low-level vision domain, it is essential to effectively harness nearby information for reconstructing LQ images. To this end, our adapter design utilizes convolutional layers to integrate nearby pixel information for restoration effectively. 
However, the direct implementation of traditional convolutional layers with a kernel size of $g$ would substantially increase parameters, growing quadratically with $g^{2}$ compared to the original fully connected (FC) layers. To mitigate this increase in parameters without significantly altering the architecture, we adopt a multi-branch structure featuring depthwise separable convolutional layers~\cite{mobilenet} and residual connections, which draw inspiration from networks with an inception-like structure~\cite{inceptionv1, inceptionv2, inceptionv3, inceptionv4, xception}. Furthermore, empirical evidence suggests that layer normalization~\cite{layernorm} and non-linear functions marginally diminish performance. Therefore, we exclude them from our adapter modules.

\subsection{Adapter Layer}
\label{subsec::adapter_layer}
The adapter layer is developed for adapting the original pre-trained module to a specific image restoration task. As demonstrated in Fig.~\ref{fig::overview}(b), each adapter layer mainly consists of two types of sub-modules: a pre-trained module and an adapter module. Since we utilize Restormer as our foundation model, the pre-trained module is the Transformer block introduced in~\cite{restormer}, which involves multi-DConv head transposed attention (MDTA), gated-DConv feed-forward network (GDFN), and two layers of layer normalization (LN). An input $h^{r}_{b, l}$ is processed through LN and MDTA, then combined with the original input to form the feature $x^{r}_{b, l}$. This feature then passes through another LN and GDFN with a residual connection to create $h'^{r}_{b, l}$. 
In specific, the process in the pre-trained module is described as follows:
\begin{align}
    &\begin{aligned}
    x^{r}_{b, l} &= h^{r}_{b, l} + \text{MDTA}( \text{LN}(h^{r}_{b, l}) ),
    \label{eq::pm_1}
    \end{aligned} \\
    &\begin{aligned}
    h'^{r}_{b, l} &= x^{r}_{b, l} + \text{GDFN}( \text{LN}(x^{r}_{b, l}) ),
    \label{eq::pm_2}
    \end{aligned}
\end{align}
where $b=\{1,2,...n,...,N\}$ denotes the index of the adapter block, $l=\{1,2,..., L_{b}\}$ represents the index of adapter layers in each adapter block. Since each restoration task has its corresponding adapter module, in Fig.~\ref{fig::overview}(b), we illustrate the process with $r=1$ to make it clear.

On the other hand, the adapter module is integrated into the pre-trained module in parallel. In adapter module $\mathcal{A}^{r}_{b, l}$, inception structure with 3x3 depthwise separable convolutional layer (DConv)~\cite{mobilenet} and 1x1 pointwise convolutional layer (PConv) are employed. Finally, $h'^{r}_{b, l}$ and $\Delta h^{r}_{b, l}$ are added together to produce the adapted output $z^{r}_{b, l}$. The overall procedure of adapter layers is formulated as follows:
\begin{align}
    &\begin{aligned}
    \Delta h^{r}_{b, l} = \mathcal{A}^{r}_{b, l}(h^{r}_{b, l}) = \text{PConv}( \text{DConv}(h^{r}_{b, l}) + \text{PConv}(h^{r}_{b, l}) ),
    \label{eq::al_1}
    \end{aligned} \\
    &\begin{aligned}
    z^{r}_{b, l} &= h'^{r}_{b, l} + \Delta h^{r}_{b, l},
    \label{eq::al_2}
    \end{aligned}
\end{align}
Integrating the adapter module into the transformer block provides the flexibility needed for the framework to adapt to different tasks, improving its overall performance and versatility.

\subsection{Training Scheme}
\label{subsec::training_scheme}
To extend the foundation model further to multiple degradations, a two-phase training strategy with a pre-training phase and a fine-tuning phase is employed.

\subsubsection{Pre-Training Phase.} In the pre-training phase, we adopt a self-supervised training strategy to enhance the model's generalizability to low-quality (LQ) input images. This strategy involves the generation of training pairs by augmenting ground truth images with various synthetic distortions, thereby creating a self-supervised learning environment. In this work, we concentrate on the influence of different pre-training schemes on the performance of downstream tasks. Our goal is to explore the shareable components across various restoration tasks. We observed that effective pre-training markedly benefits the fine-tuning process. A detailed analysis of the pre-training schemes and their effects is presented in Section~\ref{sec::analysis}.

\subsubsection{Fine-Tuning Phase.} During the pre-training phase, the foundation model is trained to extract features from a diverse array of images, effectively reducing artifacts in degraded input images. However, the foundation model may not deliver optimal performance for specific types of degradation, as its training in the pre-training phase may not be focused on those particular degradations. We incorporate adapter modules $\mathcal{A}^{r}$ into the model to address this limitation and bolster the foundation model's capabilities for specialized tasks. These adapter modules add task-specific parameters that are fine-tuned to meet the unique challenges of each task. In the fine-tuning phase, only the parameters within the adapter modules are trained, while most of the foundation model remains unchanged. This approach allows for efficient training of the AdaIR framework. After fine-tuning, a single copy of the foundation model can be stored alongside multiple lightweight adapter modules. This setup facilitates the restoration of images affected by various types of degradation without necessitating multiple large-scale models, thereby reducing both storage requirements and computational complexity.

\section{Experiments}
\label{sec::experiments}
In this section, we present experimental results and discuss their implications. We start with an introduction to our experimental setup in Section~\ref{subsec::experimental_setup}. Then, we evaluate our \modelname~with different restoration tasks in Section~\ref{subsec::main_results}. Finally, we compare different parameter-efficient tuning methods in Section~\ref{subsec::transfer_learning}.

\subsection{Experimental Setups}
\label{subsec::experimental_setup}

\subsubsection{Restoration Tasks.} We conduct experiments on four different restoration tasks, including denoising, Gaussian deblurring (GD), deraining, and super-resolution (SR). We divide LQ images of each restoration task into easy subset $E$ and hard subset $H$ to evaluate the generalizability. The datasets we use are summarized as follows:
\begin{itemize}
    \item \textbf{Denoising} In alignment with previous studies~\cite{promptir, airnet}, we conduct our experiments using a combined dataset consisting of BSD400~\cite{bsd400} and WED~\cite{wed}, totaling 4,744 images for training the model. To evaluate the denoising effectiveness of our model, we perform assessments on the BSD68~\cite{bsd68} and Urban100~\cite{urban100} datasets. Specifically, we generate noisy images through a synthetic process that adds Gaussian noise to clean images. We respectively set the noise levels $\sigma \in \{15, 25, 50\}$ to generate the easy subset, and $\sigma \in \{70, 100\}$ to generate the hard subset.
    \item \textbf{Deblurring} We have devised our experimental environment due to the limited number of studies focusing on Gaussian deblurring. We utilize the DIV2K~\cite{div2k} dataset for training, comprising 1,000 high-resolution 2K images. Out of these, 800 images are designated for the training set. The model's performance is then evaluated on the BSD68~\cite{bsd68} and Urban100~\cite{urban100} datasets. The LQ images are synthetically generated by applying blur degradations. Specifically, we use convolution with isotropic Gaussian filters to create the input images. For the Gaussian blur kernel, the kernel size $k$ and the standard deviation $\sigma$ along the two principal axes are sampled from a probability distribution to produce paired training data. We create the easy LQ images with kernel sizes $k \in \{7, 9, 11, 13, 15\}$ and standard deviations sampled from a uniform distribution $s \sim \mathcal{U}(0.2, 3.0)$. As for hard LQ images, we use kernels of sizes $k \in \{17, 19, 21\}$ and standard deviations sampled from $s \sim \mathcal{U}(3.0, 5.0)$.
    \item \textbf{Deraining} For the deraining task, our model is trained using 200 training pairs from the Rain100L~\cite{rain100l} dataset and 1,800 training pairs from the Rain100H~\cite{rain100l} dataset, corresponding to light and heavy rain streaks. To assess the effectiveness of our proposed method, evaluations are conducted on 100 testing pairs from both Rain100L~\cite{rain100l} and Rain100H~\cite{rain100l}. We categorize light and heavy rain streaks as easy subsets and hard subsets, respectively.
    \item \textbf{Super-Resolution} We use 800 images in DIV2K~\cite{div2k} to train our model to be similar to the Gaussian deblurring task. In contrast, we assess the model's performance on the Set5~\cite{set5}, Set14~\cite{set14}, B100~\cite{b100}, and Urban100~\cite{urban100} datasets. To align the image resolution with other restoration tasks and maintain a unified architecture, we initially downsample the high-resolution (HR) images to low-resolution (LR) images using bicubic interpolation. These LR images are then upscaled back to their original resolution, employing bicubic interpolation. The final outputs are merged with the input images to create the dataset for training and evaluation. We form the easy subset by sampling its scaling factor from a uniform distribution $\mathcal{U}(1, 4)$. Simultaneously, larger scaling factors of $\times 6$ and $\times 8$ are used for the hard subset.
\end{itemize}

\subsubsection{Implemantation.}
In this work, the training process is divided into two distinct phases: the pre-training phase and the fine-tuning phase. During the pre-training phase, we pre-train the foundation model using a collective dataset of the easier subsets, denoted as $E$, from all four restoration tasks to train the shareable parameters. This approach allows us to seamlessly employ any pre-trained compact foundation model, potentially eliminating the need for a dedicated pre-training phase. Subsequently, in the fine-tuning phase, we fine-tune the parameters of the task-specific adapter modules $\mathcal{A}^{r}$ using the more challenging subset $H^{r}$ for each restoration task.

\noindent During the training, the inputs are size $128 \times 128$ images, which are randomly cropped from the training set images. These images are further augmented with random horizontal and vertical flips. The batch size is set to $8$, and the AdamW optimizer~\cite{adamw} is used with the L1 loss function for training. The model undergoes training for $200$ epochs, with an initial learning rate of $2e^{-4}$. The learning rate is also adjusted according to a cosine annealing schedule~\cite{sgdr}.

\noindent Please note that the performance is evaluated in terms of peak signal-to-noise ratio (PSNR) and structural similarity index measure (SSIM). For the super-resolution (SR) task, the PSNR and SSIM evaluations are computed on the Y channel from the YCbCr color space. Conversely, for the Gaussian denoising, Gaussian deblurring, and deraining tasks, all three channels of the RGB color space are used to calculate PSNR and SSIM. The forthcoming experimental results will be presented in terms of PSNR; the SSIM results will be included in the supplementary materials.

\subsection{Validation of AdaIR}
\subsubsection{Quantitative results.}
\label{subsec::main_results}

\begin{table}[t]
\centering
\footnotesize
\resizebox{0.9\linewidth}{!}{
\begin{tabular}{l|cc|cc|cc|cc|c}
\multirow{3}{*}{Method} & \multicolumn{4}{c|}{Gaussian Denoising}                      & \multicolumn{4}{c|}{Gaussian Deblurring}                          & Deraining       \\ 
\cline{2-10}
                         & \multicolumn{2}{c|}{BSD68~\cite{bsd68}} & \multicolumn{2}{c|}{Urban100~\cite{urban100}} & \multicolumn{2}{c|}{BSD68~\cite{bsd68}}      & \multicolumn{2}{c|}{Urban100~\cite{urban100}}   & Rain100H~\cite{rain100l} \\
                         & $\sigma=70$ & $\sigma=100$  & $\sigma=70$ & $\sigma=100$     & \textit{k19s4} & \textit{k21s5} & \textit{k19s4} & \textit{k21s5} & \textit{heavy}  \\ 
\specialrule{0.8pt}{0.0ex}{0.0ex}
Restormer$_E$            & 24.22       & 18.85         & 24.44       & 18.72            & 23.33          & 22.26          & 20.02              & 19.19              & 14.48           \\
Restormer$_{E+H}$        & \bu{26.96}       & \bu{25.53}         & \bu{27.51}       & \rb{25.87}            & 27.25          & \rb{26.21}          & 25.68          & \rb{24.12}          & \bu{31.79}           \\
PromptIR$_{E+H}$         & 26.94       & 25.51        & 27.50       & \bu{25.86}            & \bu{27.26}          & \bu{26.20}          & \bu{25.72}          & \bu{24.11}          & \rb{31.91}           \\
AdaIR (Ours)             & \rb{26.98}       & \rb{25.54}         & \rb{27.54}       & \bu{25.86}            & \rb{27.29}          & 26.11          & \rb{25.73}          & 23.91          & 30.93          
\end{tabular}
}
\caption{The average PSNR (dB) over the hard subsets of the restoration tasks including Gaussian denoising, Gaussian deblurring, and deraining. The best and second-best performing results are highlighted by the \rb{red} and \bu{blue} colors, respectively.}
\label{table::main_results}
\vspace{-15pt}
\end{table}

\begin{table}[t]
\centering
\footnotesize
\resizebox{0.9\linewidth}{!}{
\begin{tabular}{l|c|c|cc|cc|cc|cc}
\multirow{2}{*}{Method} & \multirow{2}{*}{\begin{tabular}[c]{@{}c@{}}Trainable\\Param.~\end{tabular}} & \multirow{2}{*}{\begin{tabular}[c]{@{}c@{}}Training\\Time\end{tabular}} & \multicolumn{2}{c|}{Set5~\cite{set5}} & \multicolumn{2}{c|}{Set14~\cite{set14}} & \multicolumn{2}{c|}{B100~\cite{b100}} & \multicolumn{2}{c}{Urban100~\cite{urban100}}  \\
                         &                                                                            &                                                                         & $\times 6$ & $\times 8$   & $\times 6$ & $\times 8$    & $\times 6$ & $\times 8$   & $\times 6$ & $\times 8$       \\ \specialrule{0.8pt}{0.0ex}{0.0ex}
Bicubic                  & -                                                                           & -                                                                         & 24.75      & 23.09        & 22.87      & 21.64         & 22.88      & 22.05        & 19.40      & 18.48            \\ 
\specialrule{0.8pt}{0.0ex}{0.0ex}
Restormer$_E$            & 26.1M                                                                      & 43hr                                                                    & 24.20      & 23.85        & 23.11      & 22.76         & 22.67      & 23.21        & 20.95      & 20.49            \\
Restormer$_{E+H}$        & 26.1M                                                                      & 61hr                                                                    & \rb{29.22}      & \rb{26.99}        & \rb{26.61}      & \rb{25.00}         & \rb{25.97}      & \rb{24.94}        & \rb{24.13}      & \rb{22.49}            \\
PromptIR$_{E+H}$         & 35.6M                                                                      & 67hr                                                                    & 29.03      & 26.76        & \bu{26.58}      & \bu{24.86}         & \bu{25.93}      & \bu{24.93}        & \bu{24.07}      & 22.23            \\
AdaIR (Ours)             & 1.9M                                                                       & 7hr                                                                     & \bu{29.15}      & \bu{26.85}        & 26.51      & 24.83         & 25.92      & 24.87        & 24.02      & \bu{22.31}           
\end{tabular}
}
\caption{The average PSNR (dB) on Set5~\cite{set5}, Set14~\cite{set14}, B100~\cite{b100}, and Urban100~\cite{urban100} of super-resolution with scaling factor $\times 6$ and $\times 8$. Additionally, an analysis of the training time and the number of trainable parameters during training. The best and second-best performing results are highlighted by the \rb{red} and \bu{blue} colors, respectively.}
\label{table::main_sr_results}
\vspace{-20pt}
\end{table}

We compare our proposed \modelname~to existing methods for image restoration, including Restormer~\cite{restormer} and PromptIR~\cite{promptir}. Specifically, we retrain Restormer and PromptIR using LQ images from four restoration tasks simultaneously. We exploit Restormer$_E$, trained exclusively on easy LQ images $E$, as the baseline and the foundation model of  \modelname~. Furthermore, we train Restormer$_{E+H}$ and PromptIR$_{E+H}$ on a combined dataset of easy and hard LQ images.

Table \ref{table::main_results} and Table \ref{table::main_sr_results} summarize the quantitative results in terms of PSNR(dB) on Gaussian denoising, Gaussian deblurring, deraining, and super-resolution. 
Overall, Restormer$_E$ performs poorly on the hard subsets containing degradations not seen during its training. Notably, \modelname~fine-tunes Restormer$_E$ to handle the hard subsets better, significantly enhancing performance by 15.8dB PSNR on deraining task. Moreover, \modelname~achieves comparable performance with Restormer$_{E+H}$ and PromptIR$_{E+H}$. It is important to note that \modelname~ can adapt to unseen degradations, unlike Restormer and PromptIR, which are limited to predefined degradations. This comparison substantiates \modelname~'s adaptability and ability to generalize well across different types of image degradations.

As shown in Table \ref{table::main_sr_results}, the adaptation training time for \modelname~is merely 7 hours, which is relatively short compared to training Restormer and PromptIR from scratch. Meanwhile, \modelname~requires only 1.9 MB of tunable parameters, which is less than 8\% of the parameters of Restormer. This demonstrates the effectiveness and efficiency of our \modelname~framework.

\subsubsection{Qualitative results.}

\begin{figure*}[t]
    \centering
    \includegraphics[width=1\textwidth]{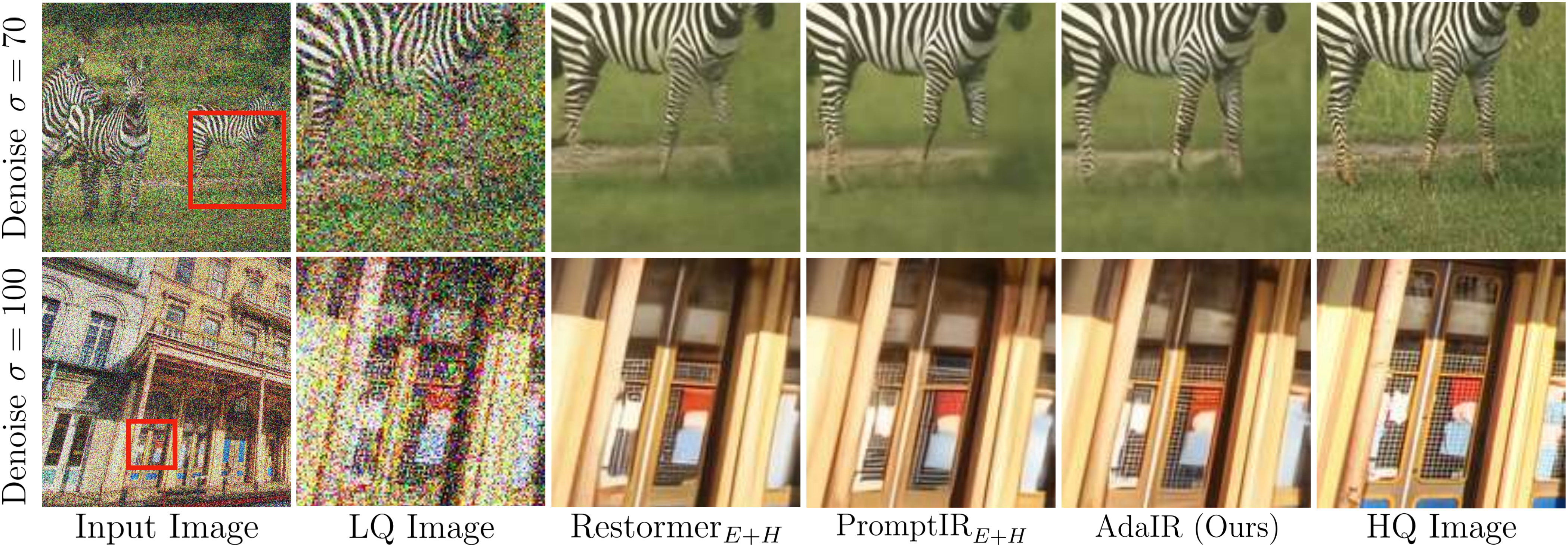}
    \caption{The qualitative results of $\text{Restormer}_{E+H}$~\cite{restormer}, $\text{PromptIR}_{E+H}$~\cite{promptir}, and our AdaIR in the denoising task with two noisy levels.}
    \label{fig::qual_denoise}
\end{figure*}
Fig.~\ref{fig::qual_denoise} compares the qualitative results of our proposed \modelname~ with the baseline method $\text{Restormer}_{E+H}$~\cite{restormer} and $\text{PromptIR}_{E+H}$~\cite{promptir}, on datasets BSD68~\cite{bsd68} and Urban100~\cite{urban100} with variant noisy level, such as $\sigma=70$ and $\sigma=100$. The visualization result of reconstructing a noisy image with $\sigma=70$ is displayed in the first row. 
The legs of the zebra are transparent or even disappear in the baseline methods. However, our proposed \modelname~restore the legs and the stripes on it. In the second row, $\text{Restormer}_{E+H}$ and $\text{PromptIR}_{E+H}$~\cite{promptir} struggle to reconstruct the mesh pattern on the image, whereas the \modelname~result demonstrate the pattern. These visualization results demonstrate the effectiveness of our method.

\begin{figure*}[t]
    \centering
    \includegraphics[width=0.9\textwidth]{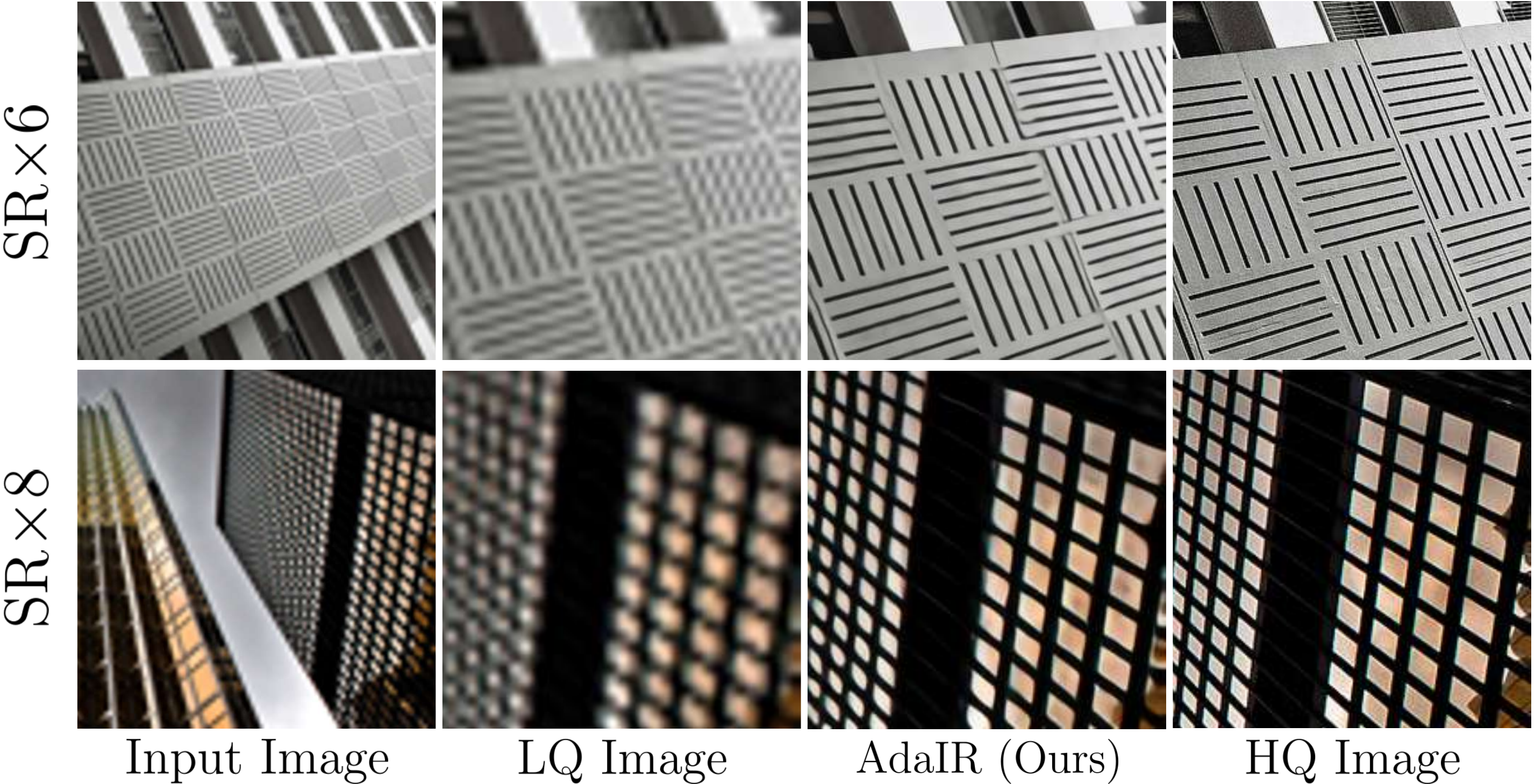}
    \caption{The qualitative results of our AdaIR in the SR task with two upscaling factors.}
    \label{fig::qual_sr}
\end{figure*}
\noindent Fig.~\ref{fig::qual_sr} demonstrates the restoration results of \modelname~on the SR task with two upscaling factors. 
The results of reconstructing the LQ images in the upscaling factor $\times6$ are illustrated in the first row. 
When the lines on the LQ image are blurry or even exhibit the wrong patterns, \ie, the horizontal lines are turned into diagonal lines. However, our method can still restore the correct pattern. On the other hand, the result of $\times8$ is shown in the second row.
Note that the rectangle windows are reconstructed by our proposed method.

\begin{figure*}[t]
    \centering
    \includegraphics[width=0.9\textwidth]{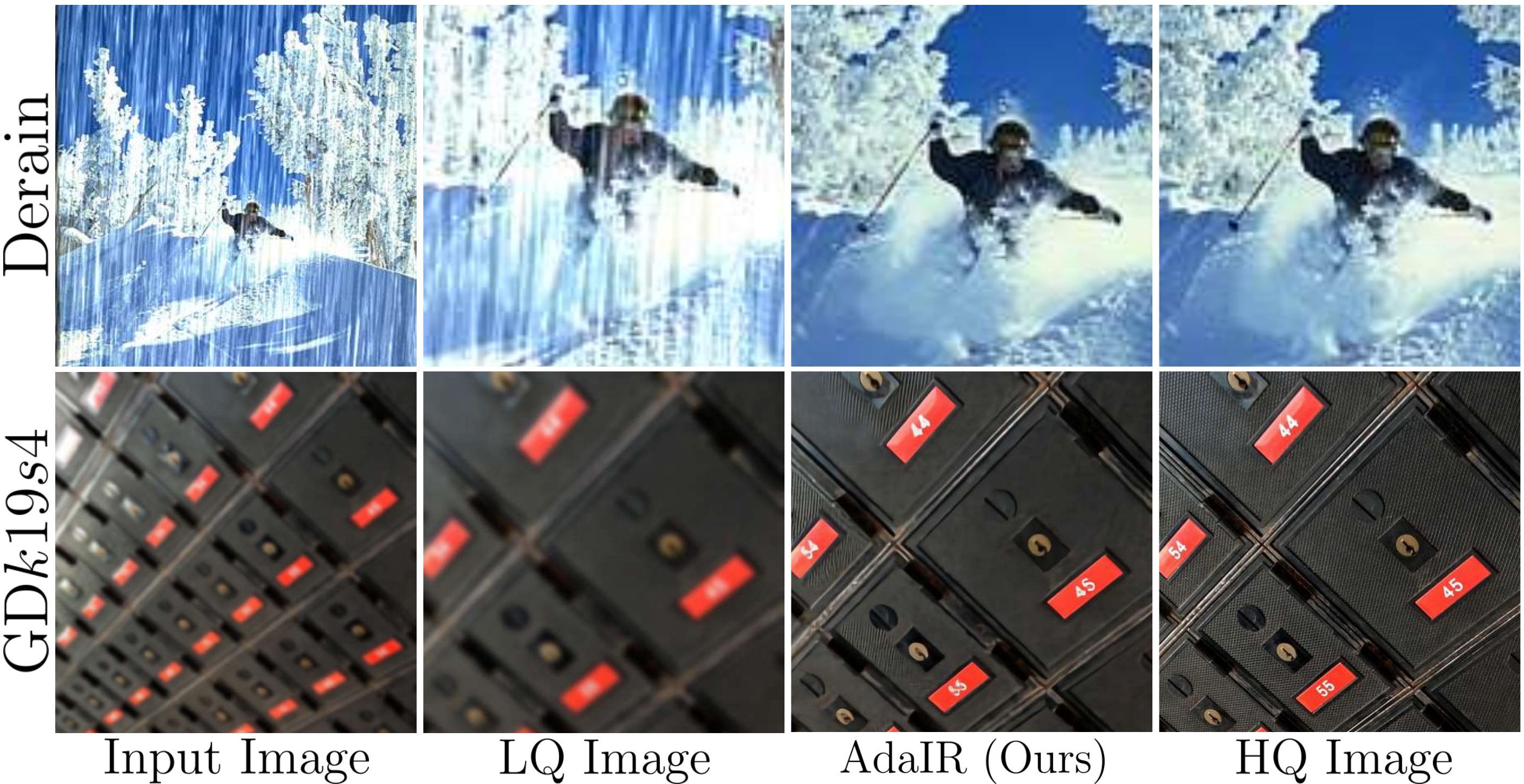}
    \caption{The qualitative results of our AdaIR in the Gaussian deblurring and deraining tasks.}
    \label{fig::qual_gd_derain}
\end{figure*}
\noindent Fig.~\ref{fig::qual_gd_derain} demonstrates the restoration results of \modelname~on the Gaussian deblurring and deraining task on the Urban100~\cite{urban100} and Rain100H~\cite{rain100l} datasets. 
As shown in the first row, rain streaks are removed even when most scenes are the same color as rain streaks. In the second row, the numbers are distorted in the LQ images, but our method can still correctly restore the numbers, such as '44' and '45'.

\subsection{Comparison of Transfer Learning Approaches}
\label{subsec::transfer_learning}
\begin{table}[t]
\centering
\footnotesize
\begin{tabular}{l|cc|cc|cc|cc}
\multirow{3}{*}{Method} & \multicolumn{4}{c|}{Gaussian Denoising}                      & \multicolumn{4}{c}{Super-Resolution} \\ 
& \multicolumn{2}{c|}{BSD68~\cite{bsd68}} & \multicolumn{2}{c|}{Urban100~\cite{urban100}} & \multicolumn{2}{c|}{B100~\cite{b100}}      & \multicolumn{2}{c}{Urban100~\cite{urban100}} \\
& $\sigma=70$ & $\sigma=100$  & $\sigma=70$ & $\sigma=100$     & $\times 6$ & $\times 8$ & $\times 6$ & $\times 8$ \\ 
\specialrule{0.8pt}{0.0ex}{0.0ex}
VPT-\textit{add}~\cite{vpt} & \bu{27.03} & 25.55 & 27.83 & 26.07 & 25.76 & 24.56 & 23.64 & 21.88 \\
AdaptFormer~\cite{adaptformer} & \rb{27.04} & \bu{25.59} & \bu{27.87} & \bu{26.21} & \bu{25.88} & \bu{24.80} & \bu{24.01} & \bu{22.20} \\
AdaIR (Ours) & \rb{27.04} & \rb{25.60} & \rb{27.88} & \rb{26.23} & \rb{25.92} & \rb{24.85} & \rb{24.10} & \rb{22.32} 
\end{tabular}
\caption{The average PSNR (dB) over the hard subsets of the restoration tasks including Gaussian denoising and super-resolution. The best and second-best performing results are highlighted by the \rb{red} and \bu{blue} colors, respectively.}
\label{table::ablation_transfer_learning}
\vspace{-25pt}
\end{table}
We compare the proposed adapter module and other parameter-efficient tuning methods, including VPT~\cite{vpt} and AdaptFormer~\cite{adaptformer}. 
It is important to note that ViT~\cite{vit} and AdaptFormer~\cite{adaptformer} were originally designed for high-level vision tasks, but we have re-implemented them for image restoration and evaluated their effectiveness. Specifically, we adopted the \textit{prepend} setting of VPT, modifying it to the \textit{add} setting as VPT-\textit{add}, where $8\times8$ learnable prompts of length $4$ are added to each hidden feature in every layer.
\noindent Table~\ref{table::ablation_transfer_learning} presents the quantitative results in terms of PSNR(dB) on Gaussian denoising and super-resolution. As shown in Table~\ref{table::ablation_transfer_learning}, our adapter-based approach outperforms the prompt-based VPT-\textit{add}, indicating that adapter-based methods may be more suitable for the image restoration domain. Compared to AdaptFormer, while our method does not exhibit a significant enhancement in the denoising task, it outperforms the super-resolution task. 
The discrepancy can be attributed to the differences in the degradation processes. Noisy images are created by adding independent Gaussian noise to each pixel, whereas LR images are produced through window-based operations, such as bicubic interpolation, which incorporate nearby pixel information. As convolutional layers excel at processing local information, our method demonstrates superior performance in the super-resolution task compared to the fully connected layers used in AdaptFormer.

\vspace{-5pt}
\section{Analysis}
\label{sec::analysis}

Shareable components play a crucial role in solving multi-task image restoration. In this study, we analyze pre-training strategies to train shareable parameters in two directions. Firstly, we investigate an inter-task scheme, which involves using different types of degradations between the pre-training and fine-tuning phases. For example, Gaussian noise may be used during pre-training, while Gaussian blur is applied during fine-tuning.
Secondly, we explore an intra-task scheme, where the same type of degradation is present in both the pre-training and fine-tuning phases, but with varying levels of severity. An instance of this would be employing Gaussian noise with a small standard deviation during pre-training and a larger standard deviation during fine-tuning. 

\subsection{Inter-Tasks Pre-Training Schemes}
\label{subsec::inter_task}
\begin{figure}[t]
    \begin{minipage}[ht]{0.55\linewidth}
        \centering
        \footnotesize
        \resizebox*{1\linewidth}{0.13\textheight}{
        \begin{tabular}{l|cc|cc|cc|c}
        \multirow{2}{*}{\backslashbox{ \scriptsize{Schemes}}{\scriptsize{Tasks}}}
        & \multicolumn{2}{c|}{Denoise} & \multicolumn{2}{c|}{SR} & \multicolumn{2}{c|}{GD} & \multicolumn{1}{c}{Derain} \\ 
        & $\sigma=70$ & $\sigma=100$ & $\times 6$ & $\times 8$ & \textit{k19s4} & \textit{k21s5} & \textit{heavy} \\ \specialrule{0.8pt}{0.0ex}{0.0ex}
        None & 25.83 & 24.54 & 23.03 & 21.76 & 26.05 & 24.73 & 25.81 \\ \specialrule{0.8pt}{0.0ex}{0.0ex}
        Denoise & \rb{27.04} & \rb{25.60} & 23.76 & 22.19 & 26.79 & 25.37 & 29.13 \\ 
        SR & 26.62 & 25.15 & \rb{24.10} & \rb{22.32} & 26.90 & 25.47 & 27.95 \\ 
        GD & 26.46 & 25.00 & 23.47 & 21.99 & \bu{27.09} & \bu{25.68} & 26.77 \\ 
        Derain & 26.47 & 25.04 & 23.33 & 21.93 & 26.53 & 25.13 & \bu{29.40} \\ \specialrule{0.8pt}{0.0ex}{0.0ex}
        All & \bu{26.98} & \bu{25.54} & \bu{24.02} & \bu{22.31} & \rb{27.29} & \rb{26.11} & \rb{30.93} 
        \end{tabular}
        }
        \vspace{-5pt}
        \captionof{table}{The average PSNR (dB) over the hard subsets of the restoration tasks including Gaussian denoising, super-resolution, Gaussian deblurring, and deraining. The best and second-best performing results are highlighted by the \rb{red} and \bu{blue} colors, respectively.}
        \label{table::inter_task}
        \vspace{-15pt}
    \end{minipage}
    \begin{minipage}[ht]{0.4\linewidth}
        \centering
        \includegraphics[width=50mm, height=0.15\textheight]{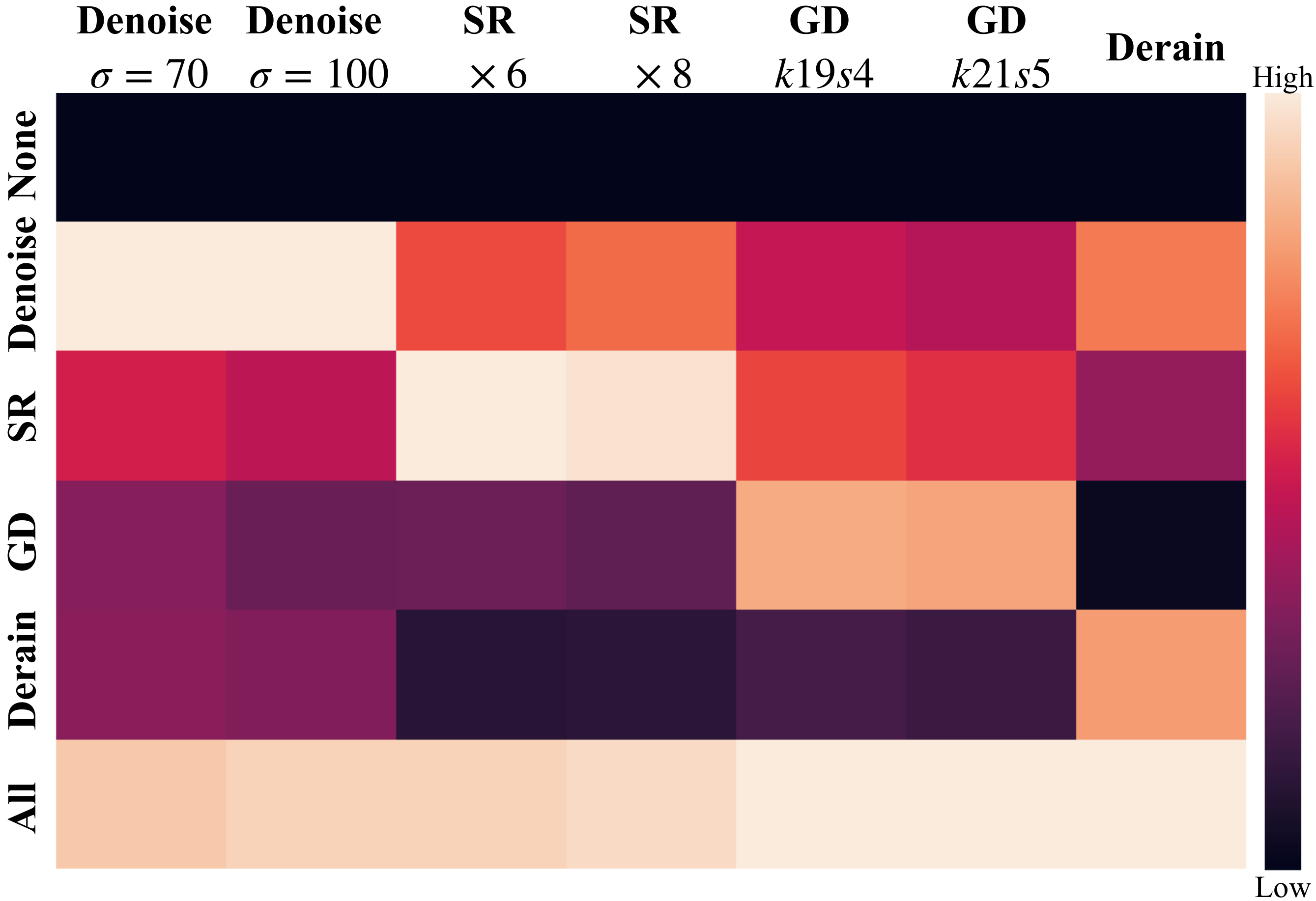}
        \captionof{figure}{Heatmap visualization of Table~\ref{table::inter_task}. Please note that the brighter the grid is, the larger the PSNR (dB) value.}
        \label{fig::heatmap}
        \vspace{-15pt}
    \end{minipage}
\end{figure}

Table~\ref{table::inter_task} presents the quantitative results of the inter-task scheme, where the pre-training scheme and the restoration tasks consider their respective types of degradation. The types of degradation in the pre-training scheme and the restoration tasks may be either the same or different. From the data in Table~\ref{table::inter_task}, it is evident that models pre-trained with any type of degradation consistently outperform those without pre-training. This observation suggests the existence of shareable components among these restoration tasks that facilitate performance in the fine-tuning phase.
For the denoising and SR tasks, the best performance is achieved when the degradation types in the pre-training scheme and the restoration tasks are the same. Conversely, for the Gaussian deblurring and deraining tasks, the best results are obtained when all types of degradations are included in the pre-training scheme. In summary, a greater correlation between the types of degradation in the pre-training scheme and the restoration tasks tends to yield better results.
\begin{figure*}[ht]
    \centering
    \includegraphics[width=1\textwidth]{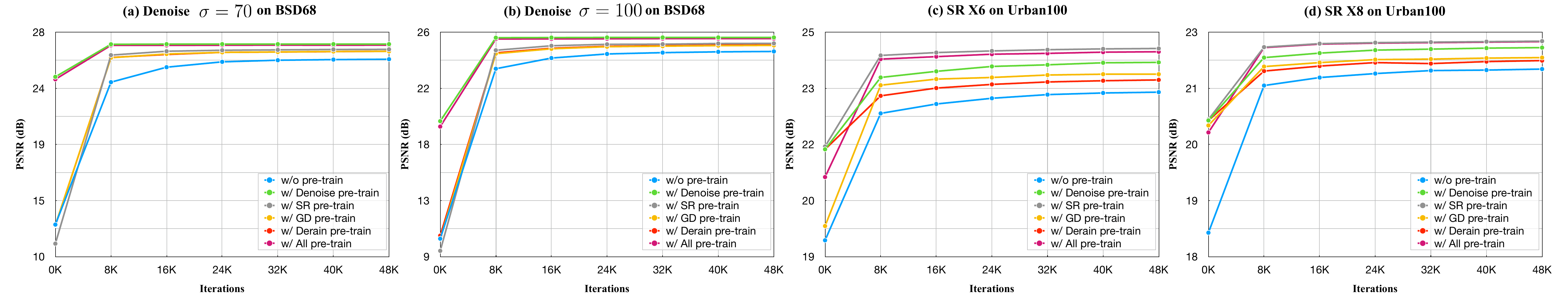}
    \caption{Performance comparison by using different pre-training schemes within the inter-tasks scope. The y-axis represents the PSNR (dB) values and the x-axis denotes the training iterations.}
    \label{fig::linechart}
\end{figure*}
\noindent Figure~\ref{fig::heatmap} visualizes the data from Table~\ref{table::inter_task} in heatmap form, clearly illustrating that pre-training is indeed beneficial for any restoration task.
To further investigate the impact of pre-training schemes on restoration tasks, we evaluate the proposed method at every $8K$ iteration interval during fine-tuning. As illustrated in Fig.\ref{fig::linechart}, we observe that when pre-training schemes more closely related to the restoration tasks are employed, the convergence speed increases. For instance, in Fig.\ref{fig::linechart}(c), the 'SR' and 'All' pre-training schemes, which are closely aligned with the restoration task SR $\times6$, provide the pre-trained foundation model with an adequate initial performance at $0$ iteration. As fine-tuning progresses, these schemes exhibit a substantial performance improvement compared to other pre-training schemes. Notably, our proposed method requires only $8K$ iterations to converge to satisfactory results.

\vspace{-5pt}
\subsection{Intra-Tasks Pre-Training Schemes}
\label{subsec::intra_task}
\begin{table}[t]
\centering
\footnotesize
\resizebox{1\linewidth}{!}{
\begin{tabular}{l|cccc|cccc}
\multirow{2}{*}{\backslashbox{ \scriptsize{Schemes}}{\scriptsize{Tasks}}}
& \multicolumn{4}{c|}{Denoise on BSD68~\cite{bsd68}} & \multicolumn{4}{c}{SR on Urban100~\cite{urban100}} \\ 
& $\sigma=55$ & $\sigma=60$ & $\sigma=70$ & $\sigma=100$ & $\times 4.5$ & $\times 5$ & $\times 6$ & $\times 8$ \\ \specialrule{0.8pt}{0.0ex}{0.0ex}
None & 26.87(\textbf{-1.20}) & 26.55(\textbf{-1.14}) & 25.83(\textbf{-1.21}) & 24.54(\textbf{-1.06}) & 24.33(\textbf{-1.59}) & 23.81(\textbf{-0.89}) & 23.03(\textbf{-0.73}) & 21.76(\textbf{-0.43}) \\ \specialrule{0.8pt}{0.0ex}{0.0ex}
Denoise & \rb{28.07} & \rb{27.69} & \rb{27.04} & \rb{25.60} & \bu{25.29}(\textbf{-0.63}) & \bu{24.70}(\textbf{-0.52}) & \bu{23.76}(\textbf{-0.34}) & \bu{22.19}(\textbf{-0.13}) \\ 
SR & \bu{27.65}(\textbf{-0.42}) & \bu{27.27}(\textbf{-0.42}) & \bu{26.62}(\textbf{-0.42}) & \bu{25.15}(\textbf{-0.45}) & \rb{25.92} & \rb{25.22} & \rb{24.10} & \rb{22.32} \\ 
GD & 27.50(\textbf{-0.57}) & 27.12(\textbf{-0.57}) & 26.46(\textbf{-0.58}) & 25.00(\textbf{-0.60}) & 25.04(\textbf{-0.88}) & 24.41(\textbf{-0.81}) & 23.47(\textbf{-0.63}) & 21.99(\textbf{-0.33}) \\ 
Derain & 27.49(\textbf{-0.58}) & 27.12(\textbf{-0.57}) & 26.47(\textbf{-0.57}) & 25.04(\textbf{-0.56}) & 24.80(\textbf{-1.12}) & 24.20(\textbf{-1.02}) & 23.33(\textbf{-0.77}) & 21.93(\textbf{-0.39}) \\
\end{tabular}
}
\caption{The average PSNR (dB) of the restoration tasks including Gaussian denoising and SR. The best and second-best performing results are highlighted by the \rb{red} and \bu{blue} colors, respectively.}
\label{table::intra_task}
\vspace{-25pt}
\end{table}

In Table~\ref{table::inter_task}, we observe that the best result for the SR $\times8$ task does not significantly outperform the results obtained when other pre-training schemes are employed. To investigate the underlying cause, we introduce the intra-task scheme. Table~\ref{table::intra_task} presents the quantitative results of this scheme. Various pre-training schemes are utilized to train the foundation model, which is then fine-tuned with a lightweight adapter module for different levels of severity within the restoration tasks.
For the denoising task, regardless of the values of $\sigma$ used in the restoration tasks, the disparity in results between the best setting and other settings remains similar. However, in the SR task, as the severity level increases from $\times4.5$ to $\times8$, the performance gaps between the best setting and other settings diminish. These observations suggest that a closer relationship between the degradation type in the pre-training scheme and the restoration tasks can lead to improved performance on the restoration tasks. The consistent performance gaps in the denoising task may be attributed to the task's inherent simplicity, resulting in a smaller distribution range across different levels of $\sigma$. Conversely, the decreasing performance gaps in the SR task as the restoration tasks become more challenging suggest that even within the same type of degradation, different levels of severity can introduce a significant domain shift.

\vspace{-5pt}
\section{Conclusion}
\label{sec::conclusion}
 In this study, we address the limitations of current methods by exploring the perspective of shareable components across multiple restoration tasks. To reach the goal, we propose the \modelname~framework, which integrates adapter modules into a common foundation model for image restoration. To enhance the generalizability of a foundation model, we employ a self-supervised strategy and diverse training data during the pre-training phase. We integrate lightweight adapter modules into the foundation model in the fine-tuning phase. These adapter modules are designed to adapt the foundation model to individual tasks. During fine-tuning, the foundation model's parameters are frozen, which allows the adapter modules to learn task-specific knowledge while preserving the general knowledge captured by the foundation model. Based on the experimental results, both quantitative and qualitative assessments demonstrate that \modelname~achieves comparable performance to current methods with fewer parameters and reduced training time. Furthermore, our comprehensive analyses of pre-training strategy assist in addressing multi-task image restoration more directly and decisively.


\bibliographystyle{splncs04}
\bibliography{citations}    

\newcommand{\IJCV}{Int. J. Computer Vision (IJCV)}\newcommand{\CVPR}{Proc. IEEE Conf. on Computer Vision and Pattern Recognition (CVPR)}\newcommand{\CVPRW}{Proc. IEEE Conf. on Computer Vision and Pattern Recognition Workshop (CVPRW)}\newcommand{\ICCV}{Proc. IEEE Int. Conf. on Computer Vision (ICCV)}\newcommand{\ICCVW}{Proc. IEEE Int. Conf. on Computer Vision Workshop (ICCVW)}\newcommand{\ECCV}{Proc. European Conf. on Computer Vision (ECCV)}\newcommand{\ECCVW}{Proc. European Conf. on Computer Vision Workshop (ECCVW)}\newcommand{\IROS}{Proc. IEEE Int. Conf. on Intelligent Robots and Systems (IROS)}\newcommand{\CoRL}{Proc. Conf. on Robot Learning (CoRL)}\newcommand{\ICRA}{Proc. IEEE Int. Conf. on Robotics and Automation (ICRA)}\newcommand{\AAAI}{Proc. AAAI Conf. on Artificial Intelligence (AAAI)}\newcommand{\IJCAI}{Proc. Int. Joint Conf. on Artificial Intelligence (IJCAI)}\newcommand{\TPAMI}{IEEE Trans. Pattern Analysis and Machine Intelligence (TPAMI)}\newcommand{\NeurIPS}{Proc. Conf. on Neural Information
  Processing Systems (NeurIPS)}\newcommand{\ICML}{Proc. Int. Conf. on Machine Learning (ICML)}\newcommand{\ICLR}{Proc. Int. Conf. on Learning Representations (ICLR)}\newcommand{\ICLRW}{Proc. Int. Conf. on Learning Representations Workshop (ICLRW)}\newcommand{\ICASSP}{Proc. IEEE Int. Conf. on Acoustics, Speech, & Signal Processing (ICASSP)}\newcommand{\BMVC}{Proc. British Machine Vision Conf. (BMVC)}\newcommand{\ACCV}{Proc. Asian Conf. on Computer Vision (ACCV)}\newcommand{\WACV}{Proc. IEEE Winter Conf. on Applications of Computer Vision (WACV)}\newcommand{\SIGGRAPH}{Special Interest Group on Compute Graphics and Interactive Techniques (SIGGRAPH)}\newcommand{\TIP}{IEEE Trans. Image Processing (TIP)}\newcommand{\ICIP}{IEEE Int. Conf. on Image Processing (ICIP)}
\begin{thebibliography}{10}
\providecommand{\url}[1]{\texttt{#1}}
\providecommand{\urlprefix}{URL }
\providecommand{\doi}[1]{https://doi.org/#1}

\bibitem{div2k}
Agustsson, E., Timofte, R.: {NTIRE} 2017 challenge on single image super-resolution: Dataset and study. In: \CVPRW. pp. 1122--1131 (2017)

\bibitem{bsd400}
Arbelaez, P., Maire, M., Fowlkes, C.C., Malik, J.: Contour detection and hierarchical image segmentation. \TPAMI pp. 898--916 (2011)

\bibitem{layernorm}
Ba, L.J., Kiros, J.R., Hinton, G.E.: Layer normalization. CoRR  (2016)

\bibitem{adapterscale}
Bapna, A., Firat, O.: Simple, scalable adaptation for neural machine translation. In: Proc. Conf. on Empirical Methods in Natural Language Processing and Int. Joint Conf. on Language Processing (EMNLP-IJCNLP). pp. 1538--1548 (2019)

\bibitem{set5}
Bevilacqua, M., Roumy, A., Guillemot, C., Alberi{-}Morel, M.: Low-complexity single-image super-resolution based on nonnegative neighbor embedding. In: \BMVC. pp. 1--10 (2012)

\bibitem{brown2020}
Brown, T.B., Mann, B., Ryder, N., Subbiah, M., Kaplan, J., Dhariwal, P., Neelakantan, A., Shyam, P., Sastry, G., Askell, A., Agarwal, S., Herbert{-}Voss, A., Krueger, G., Henighan, T., Child, R., Ramesh, A., Ziegler, D.M., Wu, J., Winter, C., Hesse, C., Chen, M., Sigler, E., Litwin, M., Gray, S., Chess, B., Clark, J., Berner, C., McCandlish, S., Radford, A., Sutskever, I., Amodei, D.: Language models are few-shot learners. In: \NeurIPS (2020)

\bibitem{dehazenet}
Cai, B., Xu, X., Jia, K., Qing, C., Tao, D.: Dehazenet: An end-to-end system for single image haze removal. \TIP pp. 5187--5198 (2016)

\bibitem{glora}
Chavan, A., Liu, Z., Gupta, D., Xing, E., Shen, Z.: One-for-all: Generalized lora for parameter-efficient fine-tuning (2023)

\bibitem{ipt}
Chen, H., Wang, Y., Guo, T., Xu, C., Deng, Y., Liu, Z., Ma, S., Xu, C., Xu, C., Gao, W.: Pre-trained image processing transformer. In: \CVPR. pp. 12294--12305 (2021)

\bibitem{hinet}
Chen, L., Lu, X., Zhang, J., Chu, X., Chen, C.: Hinet: Half instance normalization network for image restoration. In: \CVPR. pp. 182--192 (2021)

\bibitem{adaptformer}
Chen, S., Ge, C., Tong, Z., Wang, J., Song, Y., Wang, J., Luo, P.: Adaptformer: Adapting vision transformers for scalable visual recognition. In: \NeurIPS (2022)

\bibitem{hat}
Chen, X., Wang, X., Zhou, J., Dong, C.: Activating more pixels in image super-resolution transformer. CoRR  \textbf{abs/2205.04437} (2022)

\bibitem{adaptershare}
Chen, Z., Chen, B., Chen, L., Yu, K., Lou, J.: Adaptershare: Task correlation modeling with adapter differentiation. In: Proc. Conf. on Empirical Methods in Natural Language Processing (EMNLP). pp. 10645--10651 (2022)

\bibitem{xception}
Chollet, F.: Xception: Deep learning with depthwise separable convolutions. In: \CVPR. pp. 1800--1807 (2017)

\bibitem{cbm3d}
Dabov, K., Foi, A., Katkovnik, V., Egiazarian, K.O.: Color image denoising via sparse 3d collaborative filtering with grouping constraint in luminance-chrominance space. \ICIP pp. 313--316 (2007)

\bibitem{san}
Dai, T., Cai, J., Zhang, Y., Xia, S., Zhang, L.: Second-order attention network for single image super-resolution. In: \CVPR. pp. 11065--11074 (2019)

\bibitem{srcnn}
Dong, C., Loy, C.C., He, K., Tang, X.: Image super-resolution using deep convolutional networks. \TPAMI pp. 295--307 (2016)

\bibitem{msbdn}
Dong, H., Pan, J., Xiang, L., Hu, Z., Zhang, X., Wang, F., Yang, M.: Multi-scale boosted dehazing network with dense feature fusion. In: \CVPR. pp. 2154--2164 (2020)

\bibitem{fdgan}
Dong, Y., Liu, Y., Zhang, H., Chen, S., Qiao, Y.: {FD-GAN:} generative adversarial networks with fusion-discriminator for single image dehazing. In: \AAAI. pp. 10729--10736 (2020)

\bibitem{vit}
Dosovitskiy, A., Beyer, L., Kolesnikov, A., Weissenborn, D., Zhai, X., Unterthiner, T., Dehghani, M., Minderer, M., Heigold, G., Gelly, S., Uszkoreit, J., Houlsby, N.: An image is worth 16x16 words: Transformers for image recognition at scale. In: \ICLR (2021)

\bibitem{dl}
Fan, Q., Chen, D., Yuan, L., Hua, G., Yu, N., Chen, B.: A general decoupled learning framework for parameterized image operators. \TPAMI pp. 33--47 (2021)

\bibitem{derainnet}
Fu, X., Huang, J., Ding, X., Liao, Y., Paisley, J.W.: Clearing the skies: {A} deep network architecture for single-image rain removal. \TIP pp. 2944--2956 (2017)

\bibitem{lpnet}
Fu, X., Liang, B., Huang, Y., Ding, X., Paisley, J.W.: Lightweight pyramid networks for image deraining. {IEEE} Trans. Neural Networks Learn. Syst. pp. 1794--1807 (2020)

\bibitem{gao2019}
Gao, H., Tao, X., Shen, X., Jia, J.: Dynamic scene deblurring with parameter selective sharing and nested skip connections. In: \CVPR. pp. 3848--3856 (2019)

\bibitem{mam_adapter}
He, J., Zhou, C., Ma, X., Berg{-}Kirkpatrick, T., Neubig, G.: Towards a unified view of parameter-efficient transfer learning. In: \ICLR (2022)

\bibitem{kddn}
Hong, M., Xie, Y., Li, C., Qu, Y.: Distilling image dehazing with heterogeneous task imitation. In: \CVPR. pp. 3459--3468 (2020)

\bibitem{adapter}
Houlsby, N., Giurgiu, A., Jastrzebski, S., Morrone, B., de~Laroussilhe, Q., Gesmundo, A., Attariyan, M., Gelly, S.: Parameter-efficient transfer learning for {NLP}. In: \ICML. pp. 2790--2799 (2019)

\bibitem{mobilenet}
Howard, A.G., Zhu, M., Chen, B., Kalenichenko, D., Wang, W., Weyand, T., Andreetto, M., Adam, H.: Mobilenets: Efficient convolutional neural networks for mobile vision applications. CoRR  (2017)

\bibitem{lora}
Hu, E.J., Shen, Y., Wallis, P., Allen{-}Zhu, Z., Li, Y., Wang, S., Wang, L., Chen, W.: Lora: Low-rank adaptation of large language models. In: \ICLR (2022)

\bibitem{urban100}
Huang, J., Singh, A., Ahuja, N.: Single image super-resolution from transformed self-exemplars. In: \CVPR. pp. 5197--5206 (2015)

\bibitem{inceptionv2}
Ioffe, S., Szegedy, C.: Batch normalization: Accelerating deep network training by reducing internal covariate shift. In: \ICML. pp. 448--456 (2015)

\bibitem{vpt}
Jia, M., Tang, L., Chen, B., Cardie, C., Belongie, S.J., Hariharan, B., Lim, S.: Visual prompt tuning. In: \ECCV. pp. 709--727 (2022)

\bibitem{mspfn}
Jiang, K., Wang, Z., Yi, P., Chen, C., Huang, B., Luo, Y., Ma, J., Jiang, J.: Multi-scale progressive fusion network for single image deraining. In: \CVPR. pp. 8343--8352 (2020)

\bibitem{fact}
Jie, S., Deng, Z.: Fact: Factor-tuning for lightweight adaptation on vision transformer. In: \AAAI. pp. 1060--1068 (2023)

\bibitem{petl_vit}
Jie, S., Wang, H., Deng, Z.: Revisiting the parameter efficiency of adapters from the perspective of precision redundancy. In: \ICCV. pp. 17171--17180 (2023)

\bibitem{maple}
Khattak, M.U., Rasheed, H.A., Maaz, M., Khan, S.H., Khan, F.S.: Maple: Multi-modal prompt learning. In: \CVPR. pp. 19113--19122 (2023)

\bibitem{vdsr}
Kim, J., Lee, J.K., Lee, K.M.: Accurate image super-resolution using very deep convolutional networks. In: \CVPR. pp. 1646--1654 (2016)

\bibitem{deblurgan}
Kupyn, O., Budzan, V., Mykhailych, M., Mishkin, D., Matas, J.: Deblurgan: Blind motion deblurring using conditional adversarial networks. In: \CVPR. pp. 8183--8192 (2018)

\bibitem{deblurganv2}
Kupyn, O., Martyniuk, T., Wu, J., Wang, Z.: Deblurgan-v2: Deblurring (orders-of-magnitude) faster and better. In: \ICCV. pp. 8877--8886 (2019)

\bibitem{srresnet}
Ledig, C., Theis, L., Huszar, F., Caballero, J., Cunningham, A., Acosta, A., Aitken, A.P., Tejani, A., Totz, J., Wang, Z., Shi, W.: Photo-realistic single image super-resolution using a generative adversarial network. In: \CVPR. pp. 105--114 (2017)

\bibitem{prompt_tuning}
Lester, B., Al{-}Rfou, R., Constant, N.: The power of scale for parameter-efficient prompt tuning. In: Proc. Conf. on Empirical Methods in Natural Language Processing (EMNLP). pp. 3045--3059 (2021)

\bibitem{aodnet}
Li, B., Peng, X., Wang, Z., Xu, J., Feng, D.: Aod-net: All-in-one dehazing network. In: \ICCV. pp. 4780--4788 (2017)

\bibitem{airnet}
Li, B., Liu, X., Hu, P., Wu, Z., Lv, J., Peng, X.: {All-In-One Image Restoration for Unknown Corruption}. In: \CVPR. pp. 17431--17441 (2022)

\bibitem{msdinet}
Li, D., Zhang, Y., Cheung, K.C., Wang, X., Qin, H., Li, H.: Learning degradation representations for image deblurring. In: \ECCV. pp. 736--753 (2022)

\bibitem{li2020}
Li, R., Tan, R.T., Cheong, L.: All in one bad weather removal using architectural search. In: \CVPR. pp. 3172--3182 (2020)

\bibitem{rescan}
Li, X., Wu, J., Lin, Z., Liu, H., Zha, H.: Recurrent squeeze-and-excitation context aggregation net for single image deraining. In: \ECCV. pp. 262--277 (2018)

\bibitem{prefix_tuning}
Li, X.L., Liang, P.: Prefix-tuning: Optimizing continuous prompts for generation. In: Proc. Annual Meeting of the Association for Computational Linguistics and Int. Joint Conf. on Natural Language Processing (ACL/IJCNLP). pp. 4582--4597 (2021)

\bibitem{swinir}
Liang, J., Cao, J., Sun, G., Zhang, K., Van~Gool, L., Timofte, R.: Swinir: Image restoration using swin transformer. In: \ICCVW. pp. 1833--1844 (2021)

\bibitem{edsr}
Lim, B., Son, S., Kim, H., Nah, S., Lee, K.M.: Enhanced deep residual networks for single image super-resolution. In: \CVPRW. pp. 1132--1140 (2017)

\bibitem{griddehazenet}
Liu, X., Ma, Y., Shi, Z., Chen, J.: Griddehazenet: Attention-based multi-scale network for image dehazing. In: \ICCV. pp. 7313--7322 (2019)

\bibitem{sgdr}
Loshchilov, I., Hutter, F.: {SGDR:} stochastic gradient descent with warm restarts. In: \ICLR (2017)

\bibitem{adamw}
Loshchilov, I., Hutter, F.: Decoupled weight decay regularization. In: \ICLR (2019)

\bibitem{esrt}
Lu, Z., Li, J., Liu, H., Huang, C., Zhang, L., Zeng, T.: Transformer for single image super-resolution. In: \CVPRW. pp. 456--465 (2022)

\bibitem{prores}
Ma, J., Cheng, T., Wang, G., Zhang, Q., Wang, X., Zhang, L.: Prores: Exploring degradation-aware visual prompt for universal image restoration. CoRR  \textbf{abs/2306.13653} (2023)

\bibitem{wed}
Ma, K., Duanmu, Z., Wu, Q., Wang, Z., Yong, H., Li, H., Zhang, L.: Waterloo exploration database: New challenges for image quality assessment models. \TIP pp. 1004--1016 (2017)

\bibitem{compacter}
Mahabadi, R.K., Henderson, J., Ruder, S.: Compacter: Efficient low-rank hypercomplex adapter layers. In: \NeurIPS. pp. 1022--1035 (2021)

\bibitem{bsd68}
Martin, D.R., Fowlkes, C.C., Tal, D., Malik, J.: A database of human segmented natural images and its application to evaluating segmentation algorithms and measuring ecological statistics. In: \ICCV. pp. 416--425 (2001)

\bibitem{b100}
Martin, D.R., Fowlkes, C.C., Tal, D., Malik, J.: A database of human segmented natural images and its application to evaluating segmentation algorithms and measuring ecological statistics. In: \ICCV. pp. 416--425 (2001)

\bibitem{deepdeblur}
Nah, S., Kim, T.H., Lee, K.M.: Deep multi-scale convolutional neural network for dynamic scene deblurring. In: \CVPR. pp. 257--265 (2017)

\bibitem{mtrnn}
Park, D., Kang, D.U., Kim, J., Chun, S.Y.: Multi-temporal recurrent neural networks for progressive non-uniform single image deblurring with incremental temporal training. In: \ECCV. pp. 327--343 (2020)

\bibitem{adapterfusion}
Pfeiffer, J., Kamath, A., R{\"{u}}ckl{\'{e}}, A., Cho, K., Gurevych, I.: Adapterfusion: Non-destructive task composition for transfer learning. In: Proc. Conf. of the European Chapter of the Association for Computational Linguistics (EACL). pp. 487--503 (2021)

\bibitem{promptir}
Potlapalli, V., Zamir, S.W., Khan, S., Khan, F.S.: Promptir: Prompting for all-in-one blind image restoration. In: \NeurIPS (2023)

\bibitem{ffanet}
Qin, X., Wang, Z., Bai, Y., Xie, X., Jia, H.: Ffa-net: Feature fusion attention network for single image dehazing. In: \AAAI. pp. 11908--11915 (2020)

\bibitem{epdn}
Qu, Y., Chen, Y., Huang, J., Xie, Y.: Enhanced pix2pix dehazing network. In: \CVPR. pp. 8160--8168 (2019)

\bibitem{residual_adapter}
Rebuffi, S., Bilen, H., Vedaldi, A.: Learning multiple visual domains with residual adapters. In: \NeurIPS. pp. 505--516 (2017)

\bibitem{prene}
Ren, D., Zuo, W., Hu, Q., Zhu, P., Meng, D.: Progressive image deraining networks: {A} better and simpler baseline. In: \CVPR. pp. 3937--3946 (2019)

\bibitem{mscnn}
Ren, W., Liu, S., Zhang, H., Pan, J., Cao, X., Yang, M.: Single image dehazing via multi-scale convolutional neural networks. In: \ECCV. pp. 154--169 (2016)

\bibitem{rsblur}
Rim, J., Kim, G., Kim, J., Lee, J., Lee, S., Cho, S.: Realistic blur synthesis for learning image deblurring. In: \ECCV. pp. 487--503 (2022)

\bibitem{suin2020}
Suin, M., Purohit, K., Rajagopalan, A.N.: Spatially-attentive patch-hierarchical network for adaptive motion deblurring. In: \CVPR. pp. 3603--3612 (2020)

\bibitem{inceptionv4}
Szegedy, C., Ioffe, S., Vanhoucke, V., Alemi, A.A.: Inception-v4, inception-resnet and the impact of residual connections on learning. In: \AAAI. pp. 4278--4284 (2017)

\bibitem{inceptionv1}
Szegedy, C., Liu, W., Jia, Y., Sermanet, P., Reed, S.E., Anguelov, D., Erhan, D., Vanhoucke, V., Rabinovich, A.: Going deeper with convolutions. In: \CVPR. pp.~1--9 (2015)

\bibitem{inceptionv3}
Szegedy, C., Vanhoucke, V., Ioffe, S., Shlens, J., Wojna, Z.: Rethinking the inception architecture for computer vision. In: \CVPR. pp. 2818--2826 (2016)

\bibitem{srn}
Tao, X., Gao, H., Shen, X., Wang, J., Jia, J.: Scale-recurrent network for deep image deblurring. In: \CVPR. pp. 8174--8182 (2018)

\bibitem{brdnet}
Tian, C., Xu, Y., Zuo, W.: Image denoising using deep {CNN} with batch renormalization. Neural Network pp. 461--473 (2020)

\bibitem{transweather}
Valanarasu, J.M.J., Yasarla, R., Patel, V.M.: Transweather: Transformer-based restoration of images degraded by adverse weather conditions. In: \CVPR. pp. 2343--2353 (2022)

\bibitem{transformer}
Vaswani, A., Shazeer, N., Parmar, N., Uszkoreit, J., Jones, L., Gomez, A.N., Kaiser, L., Polosukhin, I.: Attention is all you need. In: \NeurIPS. pp. 5998--6008 (2017)

\bibitem{esrgan}
Wang, X., Yu, K., Wu, S., Gu, J., Liu, Y., Dong, C., Qiao, Y., Loy, C.C.: Esrgan: Enhanced super-resolution generative adversarial networks. In: \ECCVW. pp. 63--79 (2018)

\bibitem{uformer}
Wang, Z., Cun, X., Bao, J., Zhou, W., Liu, J., Li, H.: Uformer: {A} general u-shaped transformer for image restoration. In: \CVPR. pp. 17662--17672 (2022)

\bibitem{sirr}
Wei, W., Meng, D., Zhao, Q., Xu, Z., Wu, Y.: Semi-supervised transfer learning for image rain removal. In: \CVPR. pp. 3877--3886 (2019)

\bibitem{aecrnet}
Wu, H., Qu, Y., Lin, S., Zhou, J., Qiao, R., Zhang, Z., Xie, Y., Ma, L.: Contrastive learning for compact single image dehazing. In: \CVPR. pp. 10551--10560 (2021)

\bibitem{rain100l}
Yang, F., Yang, H., Fu, J., Lu, H., Guo, B.: Learning texture transformer network for image super-resolution. In: \CVPR. pp. 5790--5799 (2020)

\bibitem{vrdir}
Yang, Z., Huang, J., Chang, J., Zhou, M., Yu, H., Zhang, J., Zhao, F.: Visual recognition-driven image restoration for multiple degradation with intrinsic semantics recovery. In: \CVPR. pp. 14059--14070 (2023)

\bibitem{umrl}
Yasarla, R., Patel, V.M.: Uncertainty guided multi-scale residual learning-using a cycle spinning {CNN} for single image de-raining. In: \CVPR. pp. 8405--8414 (2019)

\bibitem{bitfit}
Zaken, E.B., Goldberg, Y., Ravfogel, S.: Bitfit: Simple parameter-efficient fine-tuning for transformer-based masked language-models. In: Proceedings of the Annual Meeting of the Association for Computational Linguistics (ACL). pp.~1--9 (2022)

\bibitem{restormer}
Zamir, S.W., Arora, A., Khan, S., Hayat, M., Khan, F.S., Yang, M.: Restormer: Efficient transformer for high-resolution image restoration. In: \CVPR. pp. 5718--5729 (2022)

\bibitem{mprnet}
Zamir, S.W., Arora, A., Khan, S., Hayat, M., Khan, F.S., Yang, M.H., Shao, L.: Multi-stage progressive image restoration. In: \CVPR. pp. 14821--14831 (2021)

\bibitem{set14}
Zeyde, R., Elad, M., Protter, M.: On single image scale-up using sparse-representations. In: Curves and Surfaces. Lecture Notes in Computer Science, vol.~6920, pp. 711--730 (2010)

\bibitem{didmdn}
Zhang, H., Patel, V.M.: Density-aware single image de-raining using a multi-stream dense network. In: \CVPR. pp. 695--704 (2018)

\bibitem{dmphn}
Zhang, H., Dai, Y., Li, H., Koniusz, P.: Deep stacked hierarchical multi-patch network for image deblurring. In: \CVPR. pp. 5978--5986 (2019)

\bibitem{idr}
Zhang, J., Huang, J., Yao, M., Yang, Z., Yu, H., Zhou, M., Zhao, F.: Ingredient-oriented multi-degradation learning for image restoration. In: \CVPR. pp. 5825--5835 (2023)

\bibitem{dncnn}
Zhang, K., Zuo, W., Chen, Y., Meng, D., Zhang, L.: Beyond a gaussian denoiser: Residual learning of deep {CNN} for image denoising. \TIP pp. 3142--3155 (2017)

\bibitem{ircnn}
Zhang, K., Zuo, W., Gu, S., Zhang, L.: Learning deep {CNN} denoiser prior for image restoration. In: \CVPR. pp. 2808--2817 (2017)

\bibitem{ffdnet}
Zhang, K., Zuo, W., Zhang, L.: Ffdnet: Toward a fast and flexible solution for cnn-based image denoising. \TIP pp. 4608--4622 (2018)

\bibitem{dbgan}
Zhang, K., Luo, W., Zhong, Y., Ma, L., Stenger, B., Liu, W., Li, H.: Deblurring by realistic blurring. In: \CVPR. pp. 2734--2743 (2020)

\bibitem{rcan}
Zhang, Y., Li, K., Li, K., Wang, L., Zhong, B., Fu, Y.: Image super-resolution using very deep residual channel attention networks. In: \ECCV. pp. 294--310 (2018)

\bibitem{rdn}
Zhang, Y., Tian, Y., Kong, Y., Zhong, B., Fu, Y.: Residual dense network for image super-resolution. In: \CVPR. pp. 2472--2481 (2018)

\bibitem{zhou2022}
Zhou, K., Yang, J., Loy, C.C., Liu, Z.: Learning to prompt for vision-language models. Int. J. Comput. Vis. (IJCV) pp. 2337--2348 (2022)

\end{thebibliography}


%
%
\end{document}